%% file: FORCE_arvix.tex
\documentclass{article}
\usepackage{iclr2027_conference,times}

\input{math_commands.tex}

\usepackage{hyperref}
\usepackage{url}
\usepackage{wrapfig}
\usepackage{algorithm}
\usepackage{algpseudocode}
\usepackage{graphicx}
\usepackage{float}
\usepackage{capt-of}
\title{FORCE: Factorized Object Relations for Counterfactual Editing in Video World Models}

\author{Suraj K. Sood \& Yugyung Lee \\
University of Missouri--Kansas City \\
Kansas City, MO, USA \\
\texttt{sks6nv@umsystem.edu, leeyu@umsystem.edu}
}

\iclrfinalcopy
\begin{document}

\maketitle
\lhead{Preprint}

\begin{abstract}
World models predict how scenes evolve, but they are difficult to
interrogate: given a model that forecasts a video, one cannot easily ask what
would have happened had a particular interaction between two objects been
different. We present Factorized Object Relations for Counterfactual Editing
(FORCE), a world model that infers pairwise interactions from raw video without
supervision and represents each as an explicit latent per object pair, so that
a relation can be removed, changed, transferred or introduced directly. Objects
are extracted by slot attention over frozen self-supervised features; a relation
encoder infers each pair latent from a short observation window, and an
interaction head turns it into a force. The latent is held fixed across the
rollout, so an edit persists rather than being recomputed, and every edit acts
on the model's own quantities with no mapping to simulator parameters. We
evaluate in a simulated environment where the true outcome of an edit is
available, scoring the change each edit induces against the change the
simulator produces under the same operation. On held-out clips, FORCE's edits made
from video agree with the simulator's response at 0.19 to 0.37 on our effect-based
score; trained and scored on exact object positions instead, the same
architecture reaches 0.57 to 0.85. The gap therefore lies not in the learned relational dynamics but in
perception: what bounds an edit is how precisely objects can be located from
video, and the precision editing needs lies just beyond what current
object-centric perception delivers. Closing that gap is a concrete target for
making counterfactual world models work from video.
\end{abstract}

\section{Introduction}
\label{sec:intro}

World models are judged by predictive accuracy, but what a model has learned
about the scene itself, beyond producing the next frames, is a separate
question.

Counterfactual queries ask it directly. If two entities are coupled, so that
one constrains the other's motion, asking how the scene would have unfolded
without that coupling requires a representation of it that can be modified on
its own; fitting the observed trajectories does not by itself provide one.

World models have followed increasingly fine factorizations of the latent
state: from a single monolithic vector in which nothing can be
isolated \citep{ha2018, hafner2020, hafner2023}, to a latent per entity
\citep{kipf2020, wu2023slotformer},
recoverable from pixels without supervision \citep{locatello2020,
seitzer2023}, and recently to a latent action per scene factor, so that one
factor can change while the others follow their own inferred actions
\citep{wang2026flam}.

Interactions were not factorized alongside them. Object-centric world models
recompute interactions at every step from the objects' recent states, by
attention over slots \citep{wu2023slotformer, wang2026flam} or message
passing over a fully connected graph \citep{kipf2020}. C-SWM's per-pair
message, for instance, is a shared function of the two objects' current
states, so an edit to it lasts one step, and it has no place for a property
of the pair that only becomes apparent over time. Models with explicit
interaction structure \citep{nakano2023stedie, feng2023daft, feng2025} use it
for prediction and control, not editing. Other work makes interactions matter
without making them addressable: C-JEPA \citep{nam2026} masks an object's
trajectory in training so that its state must be inferred from the others,
evaluated by counterfactual question answering on CLEVRER \citep{yi2020},
and another line alters the observed initial condition and forecasts from it
\citep{baradel2020, janny2022}. Neither changes the model's own
representation of an interaction.

On ground-truth state, relational structure is well studied: Interaction
Networks \citep{battaglia2016} propagate dynamics over a given graph, NRI
\citep{kipf2018} infers the graph, establishing per-pair latents as the unit
of relational representation, and NIIP \citep{comas2023} swaps per-pair
energy functions between separately trained models.
V-CDN \citep{li2020vcdn} carries per-pair latents to video: it infers a type
and parameters for each directed pair of keypoints, holds them fixed over an
episode, and predicts counterfactual futures when they change. Its edits,
however, pass through a mapping to simulator parameters fitted with ground
truth, and are scored by trajectory error, which mixes the edit's effect with
ordinary prediction error.

We introduce FORCE, which carries the factorization to interactions and makes
the relation itself the object of the edit. A relation encoder infers a latent
$\mathbf{m}_{ij}$ for each ordered pair from a short observation window, and the
latent is held fixed for the rest of the rollout. Unlike a per-step message, it is read from motion over the window,
so it can carry properties of the pair that only become apparent over time,
and an edit to it persists rather than being recomputed. An interaction head
computes the force at each step from $\mathbf{m}_{ij}$ and the current
configuration, so the relation is addressable while its effect varies; editing it gives the
model's prediction for a world in which that relation differs. We consider four edits (removing a relation, changing its
parameters, transferring one between pairs, and introducing one where none
existed), all acting only on the model's own quantities, with no knowledge of
the simulator's parameterization. Holding $\mathbf{m}_{ij}$
fixed assumes relations persist; we state this explicitly and return to it in
the limitations.

Entities come from slot attention over frozen self-supervised features
\citep{locatello2020, oquab2023}; relations are inferred without adjacency
labels or simulator state at any stage of training, and a label-free rule
selects which pairs are related. Because the environment is simulated, the
true counterfactual is available; following treatment-effect evaluation
\citep{hill2011, shalit2017}, we score the change an edit induces, the difference between the model's edited and unedited
rollouts, against the change the simulator produces under the same edit.
Scoring the effect rather than the outcome matters here: for three of the four
edits, the rollout's own error is four to eleven times the change the edit
truly induces, so outcome error barely registers whether an edit was applied.

From video FORCE performs considerably worse than from exact entity
states, and characterising that gap is our second contribution. Retraining
on ground-truth positions corrupted with controlled noise traces edit
accuracy against perceptual error and shows a sharp threshold near 1.7 pixels, beyond which the learned dynamics
predict worse than assuming the objects move freely.
Substituting exact values one at a time locates the shortfall: the largest
share comes from the rollout's starting state, particularly its velocity,
which is hard to estimate from noisy measurements
\citep{chartrand2011}.

This is not specific to our architecture. Object-centric world models are
assessed on segmentation quality and rollout fidelity (Section~\ref{sec:rw-oc}), and
Filtered-CoPhy reports, in an appendix, keypoint distance to the
true object centres \citep{janny2022}, but we know of no learned counterfactual
world model on video that traces edit accuracy against position error. Physion suggests that extracting physical representations is the main
bottleneck for physical prediction from video \citep{bear2021}; for relational editing, we find a
narrower one: positional precision below two pixels.

Our contributions:

\begin{itemize}
\item \textbf{FORCE, a relationally factored world model with editable
relations}: a relation encoder infers one latent per ordered pair from video
without simulator state, held fixed so that edits persist, and an interaction
head turns it into a force. Removing, changing, transferring and introducing
a relation all act on the model's own quantities and require no mapping from
its latent space to the simulator's parameters.
\item \textbf{Effect-based evaluation of relational edits} against the
simulator's change under the same operation, separating the edit from
ordinary prediction error.
\item \textbf{A measurement of the precision relational editing requires}:
under controlled noise, a sharp threshold near 1.7 pixels, against
3.8 pixels of error in slot-based perception.
\item \textbf{A decomposition of the video-to-state gap}, identifying the
initial velocity, not the learned dynamics, as the dominant source of error.
\end{itemize}

\section{Related Work}

\subsection{Object-centric perception and world models}
\label{sec:rw-oc}
Slot Attention binds latent slots to scene entities without supervision
\citep{locatello2020}, extended to video \citep{kipf2022savi}, to real images
through self-supervised feature reconstruction \citep{seitzer2023}, and with
explicit per-slot position and scale \citep{biza2023}; keypoint and particle
models instead represent each object as a point with local features
\citep{kulkarni2019transporter, daniel2022dlp}. Evaluation is by segmentation
quality (FG-ARI, mBO), including the closest study relating it to planning
\citep{nazirjonov2026}; how precisely a slot localizes its object is rarely
reported, though human annotators tracking points disagree by only about 1.5
pixels at $256\times256$ \citep{doersch2022tapvid}, close to the precision
editing requires. We read positions from the attention maps of a
DINOSAUR-style encoder over frozen DINOv2 features \citep{oquab2023}; their
precision, not segmentation quality, limits editing.

Beyond the models discussed in Section~\ref{sec:intro} \citep{kipf2020, wu2023slotformer,
wang2026flam, nam2026}, object-centric world models improve slot decoding with
diffusion \citep{wu2023slotdiffusion} and extend to control through latent
dynamics \citep{mosbach2025sold}, latent actions
\citep{villarcorrales2025playslot, klepach2025} and particle dynamics
\citep{daniel2026lpwm}; earlier work discovered objects and their interactions
jointly from pixels \citep{vansteenkiste2018rnem}. Interaction structure is made
explicit through interaction-relevant object features \citep{nakano2023stedie},
an interaction graph over object attributes \citep{feng2023daft} or interaction
primitives \citep{feng2025}; RELATE edits structured object latents
\citep{ehrhardt2020relate}, and O2P2 applies a pairwise interaction function to
object representations \citep{janner2019o2p2}. All are assessed on rollout
fidelity, planning or control, and in these models interactions are recomputed
from the current object states, so an edit to them does not persist.
Relational-inference models hold a per-pair latent fixed
(Section~\ref{sec:rw-ri}), but over given states or keypoints, not inside a
world model built on slots.

\subsection{Relational inference and learned physics}
\label{sec:rw-ri}
Interaction Networks \citep{battaglia2016}, the Neural Physics Engine
\citep{chang2017npe} and graph network simulators
\citep{sanchezgonzalez2020gns} learn dynamics over a given graph, and
Visual Interaction Networks infer the state from pixels first
\citep{watters2017}. NRI infers the graph from trajectories as
discrete edge types \citep{kipf2018}, with meta-learned \citep{alet2019},
factorized \citep{webb2019fnri}, dynamic \citep{graber2020} and amortized
causal \citep{lowe2022acd} extensions; NIIP represents each interaction as an
energy potential \citep{comas2023}. From video, V-CDN infers per-edge latents
over unsupervised keypoints and intervenes through a mapping to simulator
parameters \citep{li2020vcdn}, while other work recovers object properties by
pairing perception with a physics model \citep{wu2015galileo, jaques2020paig,
ding2021vrdp} or identifies spring and other physical parameters from video
through differentiable simulation \citep{kandukuri2022, castanedagarcia2025,
zhong2024springgaus, jiang2025phystwin}. Graph recovery depends on node degree
\citep{matta2019degree}, and causal discovery degrades under measurement error
\citep{zhang2017measurement}; we restrict our study to sparse graphs, and our
precision sweep measures the effect of measurement error directly. FORCE's
relation latent plays the role of NRI's edge and is likewise held over the
rollout, but is continuous, inferred from slot trajectories, and applied
through an interaction head whose forces are equal and opposite; edits write
the coefficients the head reads from that latent, with no mapping to simulator
parameters.

\subsection{Counterfactual evaluation of physical scenes}
Counterfactual physical reasoning has been posed as question answering
\citep{yi2020, chen2022comphy}, as forecasting from a modified initial
condition against simulated ground truth \citep{baradel2020, janny2022}, and
as physical prediction from vision \citep{bear2021, tung2023physionpp};
elsewhere a masked predictor's input is perturbed \citep{venkatesh2024cwm},
counterfactual dynamics teach object interactions \citep{choi2024coil}, and
latent causal factors are identified from image sequences with known
intervention targets \citep{lippe2022citris}. In each, the intervention acts
on the input or on single latent factors, such as an object's friction
\citep{ye2018iip}, or the evaluation asks about an outcome; among learned models,
only V-CDN edits an inferred pairwise interaction and scores the result against
a simulator. Counterfactual predictions are also fragile to small
errors in state or parameters \citep{aalaila2025chaos}. Throughout, a
relation is a persistent property of an ordered pair, as a spring has;
interactions arising only on contact are a different object
(Section~\ref{sec:limitations}). Concurrently, Do-JEPA \citep{resani2026dojepa} trains slot world models on
paired simulator rollouts that differ in the agent's action; FORCE instead
edits relations and uses no simulator state in training.

\section{Method}
\label{sec:method}

\begin{figure}[!h]
\centering
\includegraphics[width=\linewidth,trim=0 7.5 0 3.5,clip]{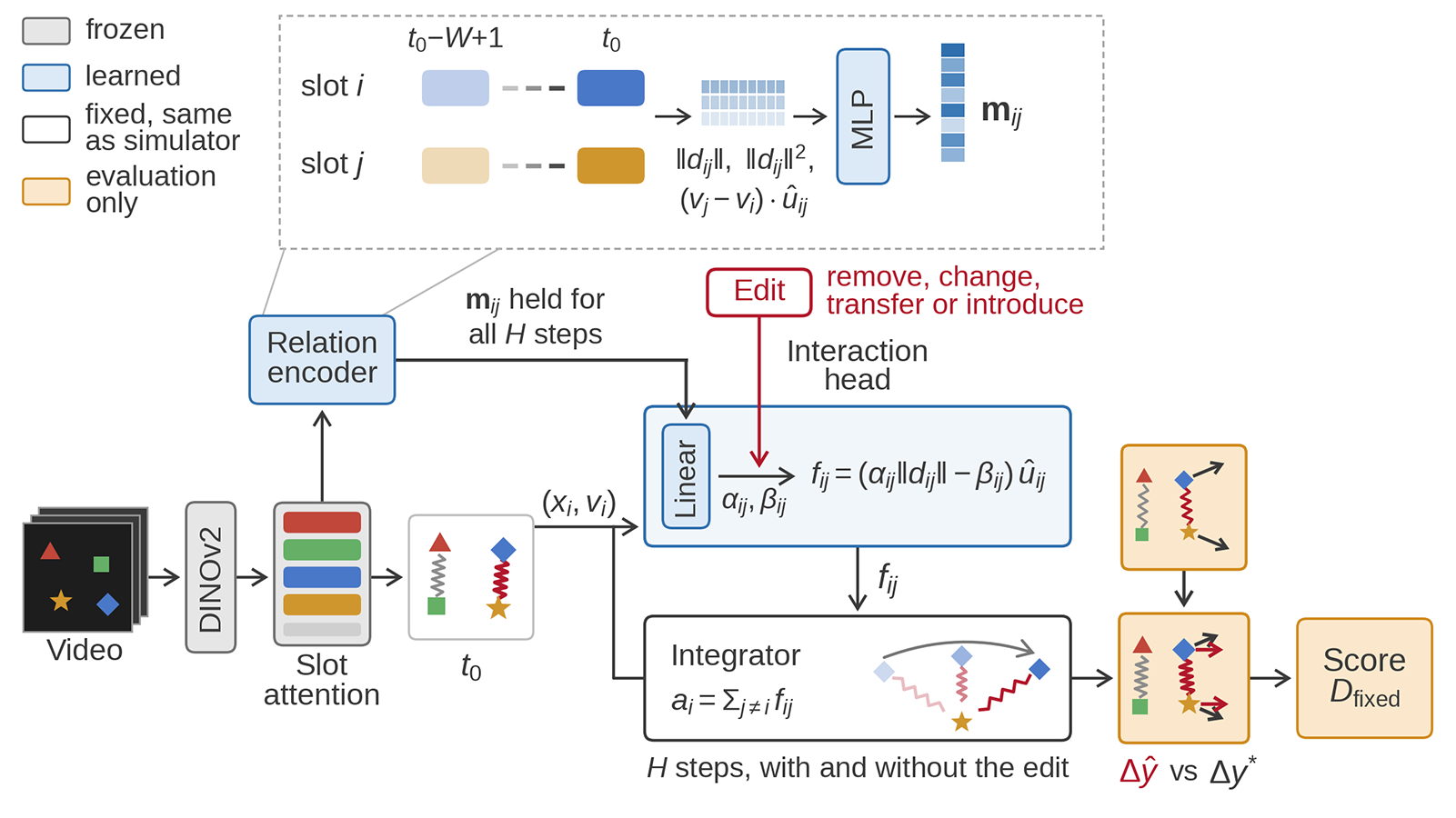}%
\caption{Overview of FORCE. Slot attention over frozen DINOv2 features yields
entities and their positions; the relation encoder reads each pair's motion over
the observation window and infers a relation latent, held fixed for the rollout;
the interaction head turns each latent, with the current configuration, into a
force; and a fixed integrator that follows the simulator's update advances the
state. An edit writes the coefficients $\alpha_{ij}, \beta_{ij}$ the interaction
head reads from a pair's relation latent, and its effect $\Delta\hat{y}$ is
scored against the simulator's $\Delta y^{*}$ under the same edit.}
\label{fig:pipeline}
\end{figure}

FORCE models a scene as a set of entities whose pairwise interactions are
represented explicitly, inferred once from a short observation window and held
fixed while the scene is rolled forward. Editing an interaction means
overwriting the quantities the model inferred for that pair. The pipeline
has four parts, shown in Figure~\ref{fig:pipeline}. Slot attention
over frozen self-supervised features yields entities and their positions; a
\emph{relation encoder} infers one latent per ordered pair from a short window
of motion; an \emph{interaction head} maps that latent to a force; and a fixed
integrator advances the state. The slots, relation encoder and interaction head
are learned from video, the integrator is fixed and follows the simulator's
update, and an edit changes only the coefficients the head reads from one
pair's relation latent. A diffusion decoder renders frames for figures only
(Appendix~\ref{app:decoder}).

\subsection{Setup}
A clip is a sequence of RGB frames at $224\times224$ containing $N$ entities.
The model observes a window of $W=16$ frames ending at the anchor $t_0$, frames
$t_0-W+1,\dots,t_0$, and predicts the next $H$.
Entity states are positions $x_i$ in the unit square and velocities $v_i$, in the same
units the simulator uses, so that predicted and true trajectories are directly
comparable. No simulator state of any kind is used during training; it appears
only at evaluation, as the answer key.

\subsection{Entities}
Frames are encoded by a frozen DINOv2 backbone \citep{oquab2023}, and slot
attention \citep{locatello2020} binds $K$ slots to the resulting patch features,
following the feature-reconstruction formulation of \citet{seitzer2023}. Slots
are trained in a first stage and frozen thereafter.

Each slot yields an attention map over the $16\times16$ patch grid. We normalise
that map and take its centroid as the entity's position, rescaled to $[0,1]$.
Velocity is estimated by optical flow \citep{farneback2003} between consecutive
frames, pooled under the slot's own attention map. One slot is identified as background, without
supervision, as the slot with the largest attention mass at the start of the
window; the remaining slots are the entities, taken in index order.

\subsection{Relations}
\label{sec:relations}
For each ordered pair $(i,j)$ we form per-frame features from the window,
\begin{equation}
\phi_{ij}^{(t)} = \big[\; \lVert d_{ij}^{(t)} \rVert,\;
\lVert d_{ij}^{(t)} \rVert^{2},\;
(v_j^{(t)} - v_i^{(t)}) \cdot \hat{u}_{ij}^{(t)} \;\big],
\qquad d_{ij} = x_j - x_i ,
\end{equation}
with $\hat{u}_{ij}$ the unit vector from $i$ to $j$. These are concatenated over
the window and mapped by the \emph{relation encoder}, an MLP, to an
eight-dimensional relation latent $\mathbf{m}_{ij}$, symmetrised as
$\mathbf{m} \leftarrow \tfrac{1}{2}(\mathbf{m} + \mathbf{m}^{\top})$.

Two properties matter: $\mathbf{m}_{ij}$ is read from motion across the whole
window, so it can carry a property of the pair visible only over time, and it is
held fixed for the entire rollout, so an edit persists instead of being
overwritten by recomputation at the next step.

\subsection{Dynamics}
The \emph{interaction head} maps the relation latent to a force. In the form
used here it is a linear layer,
$(\alpha_{ij}, \beta_{ij}) = U \mathbf{m}_{ij} + c$, followed by a central
force
\begin{equation}
f_{ij} = \big(\alpha_{ij} \lVert d_{ij} \rVert - \beta_{ij}\big)\, \hat{u}_{ij},
\qquad
a_i = \sum_{j \neq i} f_{ij},
\end{equation}
which is equal and opposite by the symmetry of $\mathbf{m}$. The state is then
advanced by a fixed integrator that implements the simulator's own update rule:
forces are accumulated, velocity is updated and capped, position is advanced,
and wall contacts reflect the velocity component normal to the wall. The model
and the simulator follow this rule in training, scoring and the reference
trajectories, so that no discrepancy between the model's and the simulator's
physics can pass for a modelling result.

\subsection{Editing a relation}
A counterfactual question asks what the scene would have done had one
interaction been different; answering it needs a variable that represents that
interaction and survives long enough for a change to matter, which holding
$\mathbf{m}_{ij}$ provides (Section~\ref{sec:relations}). An edit is a write to
the quantities the model inferred for a chosen pair, after which the scene is
rolled forward unchanged. Edits act on the coefficients $\alpha_{ij}, \beta_{ij}$
the interaction head reads from $\mathbf{m}_{ij}$, after the selection described
below. For a target pair
$(i,j)$ and, where needed, a donor pair $(k,l)$, the four operations are:

\begin{itemize}
\item Remove. $\alpha_{ij} = \beta_{ij} = 0$: the pair exerts no force.
\item Change. $\beta_{ij} \leftarrow \beta_{ij} + \delta\,\alpha_{ij}$, which
shifts the pair's rest distance by $\delta$ as the model itself estimates it,
since $\hat r_{0} = \beta / \alpha$.
\item Transfer. $\beta_{ij} \leftarrow \alpha_{ij}\,(\beta_{kl}/\alpha_{kl})$:
the target pair takes a donor pair's rest distance, read from the donor's own
coefficients.
\item Introduce. $(\alpha_{ij}, \beta_{ij}) \leftarrow (\alpha_{kl}, \beta_{kl})$
for an unrelated pair, copying the donor's coefficients.
\end{itemize}

Every written value is a quantity the model produced, or a fixed constant. No
mapping from the latent space to the simulator's parameterisation is fitted, and
no ground-truth quantity enters the edit. Ground truth enters only in specifying
\emph{which} pair the request concerns, which is a property of the question
rather than of the method: if the model did not select that pair, its
coefficients are zero and the edit has no effect, which the metric records.

A label-free rule decides which pairs are treated as related. Pairs are ranked
by $\sqrt{\alpha^2 + \beta^2}$ and split by an Otsu threshold
\citep{otsu1979};
the retained mask multiplies $\alpha$ and $\beta$ before the rollout. The rule
reads no adjacency. Selection is applied at evaluation only; training sums
forces over all pairs.

\subsection{Scoring an edit}
Because the environment is simulated, the trajectory the world would have taken
under the same edit is available. We therefore score the \emph{change} an edit
induces rather than the state it produces. With $\hat y$ the model's rollout and
$y^{*}$ the simulator's,
\begin{equation}
\Delta \hat y = \hat y_{\text{edit}} - \hat y_{\text{factual}},
\qquad
\Delta y^{*} = y^{*}_{\text{edit}} - y^{*}_{\text{factual}},
\end{equation}
both evaluated at the end of the rollout, per entity. The metric is the mean
cosine between them over entries whose true response is non-zero:
\begin{equation}
D_{\text{fixed}} = \underbrace{\Pr\!\big[\lVert \Delta \hat y \rVert > 0\big]}_{\text{coverage}}
\times
\underbrace{\operatorname{mean}\big[\cos(\Delta \hat y, \Delta y^{*})\big]}_{\text{quality}} .
\end{equation}
Entries where the model produces no response contribute zero. Scoring the
difference between potential outcomes rather than the outcomes follows
treatment-effect evaluation \citep{hill2011, shalit2017}, and matters here
because the rollout's own error is 1.5 to 11 times the true effect of an edit.
Coverage is in practice the fraction of requests for which the model selected
the requested pair. We also report a variant of the metric
restricted to the two edited entities, which separates the response of the pair
itself from its propagation to entities the edit never touched.

Reference trajectories branch the simulator from its stored state at the start
of the rollout, so the factual and edited references differ only by the edit, and
the null edit reproduces the factual trajectory exactly. Entities are matched to
the simulator's objects by position only to compare changes object by object,
which permutes indices and cannot change a score.

\subsection{Training}
Training has two stages and no simulator state enters either.

\emph{Perception.} Slot attention is trained over frozen DINOv2 features with a
feature-reconstruction objective and a short prediction rollout, then frozen;
the full objective, with its terms and weights, is given in the appendix.
Entity positions and velocities are cached once for the whole dataset.

\emph{Relations.} The relation encoder and the interaction head are trained
jointly from scratch on the cached streams, by minimising position error
over a short rollout,
$\mathcal{L} = \tfrac{1}{H}\sum_{h} \lVert \hat x^{(h)} - x^{*(h)} \rVert^{2}$,
where $x^{*}$ is the cached observation, not simulator state. The rollout is
free-running from the first step of training: the model's own predicted state is
fed back through the integrator, the observations serve only as targets, and the
relation latent is inferred once per window and held, exactly as at evaluation. A guard asserts
throughout training that no simulator quantity is read.

Training and evaluation differ in horizon and in position and velocity estimates
(Appendix~\ref{app:relations}).

\section{Experiments}
\label{sec:experiments}

In this section we evaluate whether FORCE's relations, inferred from video, can
be edited so that its response matches the simulator's, isolate what that response
depends on, and measure what limits it.

\subsection{Experimental setup}

Because no prior method edits a relation
latent directly from video, the comparisons are controls rather than competing
systems, and each carries a stated pass condition. The edit-direction prior
pushes the named pair the direction the edit would plausibly move it, with no
model; an edit that does not beat it is not using the relation. The wrong-pair
control performs the edit on a different pair and scores against the truth for
the original, so a pair-specific response gives zero. The re-derived ablation
recomputes the relation latent from the predicted state at every step instead of
holding it, removing the mechanism the method rests on. Distance-only selection
replaces the learned selection with a single statistic of the pair distance,
chosen and oriented on a training split, and asks whether the learned latent
selects better than geometry. We also report an oracle-selection row that
substitutes the true adjacency; any number that substitutes a simulator quantity
is a diagnostic and is labelled as one wherever it appears.

Four objects of radius $0.08$ (17.9 px at $224\times224$) move in a unit box
under two springs, drawn independently per simulation and fixed for its duration;
each is rendered as a distinct solid shape, because frozen DINOv2 features tell
objects apart mainly by form (Appendix~\ref{app:env}). Rest lengths are
randomised so that no single separation marks a linked pair: on exact positions,
a pair's mean separation predicts whether it is linked at AUC $0.51$, against
$0.68$ when every rest length is fixed. A linked pair's separation still stays
near its own rest length, which the distance control in
Section~\ref{sec:controls} exploits. Entities come from
slot attention with five slots over frozen DINOv2-S/14 features
\citep{locatello2020, oquab2023}, trained once and frozen
(Appendix~\ref{app:stage0}); a cell of the feature grid is 14 px, so the
precision editing needs, measured in Section~\ref{sec:cliff}, is about an eighth
of a cell.

\begin{table}[!h]
\centering
\caption{Editing performance. $D_{\text{fixed}} = \text{coverage} \times
\text{quality}$. Development and held-out are five seeds, and the held-out
column is the headline; coverage and quality are on development. The state
column is the same architecture trained and scored on true positions, one
seed, and is a diagnostic.}
\label{tab:main}
\vspace{4pt}
\small
\begin{tabular}{lccccc}
\hline
edit & development & held-out & coverage & quality & state \\
\hline
Remove    & $0.1954 \pm 0.0051$ & $0.1888 \pm 0.0205$ & 0.817 & 0.239 & 0.6275 \\
Change    & $0.3650 \pm 0.0080$ & $0.3720 \pm 0.0224$ & 0.817 & 0.447 & 0.6295 \\
Transfer  & $0.1456 \pm 0.0123$ & $0.2001 \pm 0.0226$ & 0.817 & 0.179 & 0.5712 \\
Introduce & $0.3198 \pm 0.0021$ & $0.2860 \pm 0.0161$ & 0.901 & 0.355 & 0.8482 \\
\hline
\end{tabular}
\end{table}

The relation encoder and interaction
head are trained with an observation window of $W=16$ frames and a training
horizon of $H=5$ (Appendix~\ref{app:relations}); the interaction head sees an all-ones adjacency during training
and must silence unrelated pairs itself, and selection is applied only at
evaluation. Everything is split by simulation and never by clip, because
neighbouring clips overlap heavily: 280 training, 40 validation and 80 test
simulations. Development fixtures draw 200 clips from test simulations, with
anchors restricted to the range in which the simulator's counterfactual replay
is exact. The scoring horizon $H=10$ was selected on the development fixture from a sweep
of horizons (Appendix~\ref{sec:horizon}). We additionally hold out 800 clips from
80 further simulations, generated separately and kept out of development
entirely, and score them at the chosen
horizon under a frozen recipe; this is the headline number. Every number is five
seeds unless stated, reported as mean $\pm$ SE for development and mean $\pm$
SD across seeds for the held-out clips, with bootstrap intervals resampled by
simulation rather than by clip, and all arms of a comparison share a training
budget, an evaluation fixture, a selection rule and a scoring horizon.

\subsection{Editing performance}
\label{sec:main}

\begin{figure}[!h]
\centering
\includegraphics[width=\linewidth,trim=5 9.5 0 0,clip]{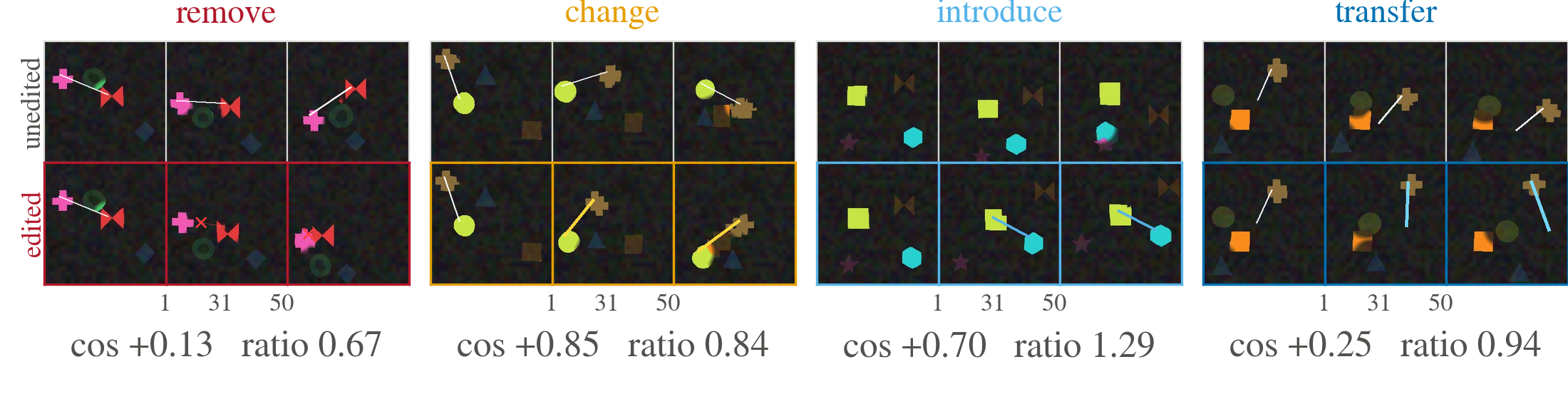}%
\caption{The four edits rendered by the diffusion decoder, with the unedited
rollout above the edited one over 50 steps. Each clip is the one with the
largest true response for that edit, chosen without looking at the model's
predictions. The frames are for illustration; every reported number is
computed on predicted states.}
\label{fig:edits}
\end{figure}

Table~\ref{tab:main} gives $D_{\text{fixed}}$ for the four edits. All four
produce a response aligned with the simulator's (Figure~\ref{fig:edits}), all
four sit well below the state-space reference, and recovering the relation graph
from the coefficient norm reads AUC $0.8124$ on development and $0.8264$ on the
held-out clips. The held-out values track development closely: removal and
change differ by less than $0.01$, introduction falls by $0.034$, and transfer
rises by $0.055$ because the held-out clips happen to request larger transfers,
which the model handles better.

\subsection{Ablations and controls}
\label{sec:controls}

\begin{table}[!h]
\centering
\caption{Controls and ablations, five seeds. The oracle row substitutes true
adjacency and is a diagnostic.}
\label{tab:controls}
\vspace{4pt}
\small
\begin{tabular}{lcccc}
\hline
 & Remove & Change & Transfer & Introduce \\
\hline
FORCE                      & 0.1954 & 0.3650 & 0.1456 & 0.3198 \\
edit-direction prior       & $-0.0209$ & 0.2926 & 0.0253 & $-0.0044$ \\
wrong pair                 & $+0.0134$ & $-0.0503$ & $+0.0022$ & $-0.0255$ \\
relation re-derived        & 0.0399 & 0.3042 & 0.1260 & 0.2285 \\
distance-only selection    & 0.2030 & 0.4156 & 0.1605 & 0.3184 \\
oracle selection (diag.)   & 0.2481 & 0.4748 & 0.1841 & 0.3316 \\
\hline
\end{tabular}
\end{table}

Table~\ref{tab:controls} gives the four comparisons. All four edits clear the
edit-direction prior, and the wrong-pair control stays small for every edit, at
most about a seventh of the edit's own value, so each response is specific to
the pair named. Holding the
relation is what keeps an edit alive over a rollout. If the latent is re-derived
from the predicted state at every step instead, it drifts back towards its
unedited value and the edit fades: at $H=10$ removal loses 80\% of its value and
the other three edits lose 13 to 29\%, and at 21 steps removal is
lost entirely and the others lose about two thirds. The longer the prediction, the more holding
matters, and for removal it is essential at either length.

A distance statistic selects pairs at least as well as FORCE's relation latent.
It matches FORCE on introduction and is ahead on the other three edits, and on
the development clips it recovers the graph at AUC $0.845$ against FORCE's
$0.812$. The statistic
is $|\bar d_{ij} - \tilde d|$, the deviation of a pair's mean separation from the
population median, chosen and oriented on a training split. We therefore do not
claim relation discovery from video as a contribution. The mechanism is worth
stating: a spring holds its pair near its rest length, so a linked pair's
separation sits near a characteristic value and stays there, and any method that
watches separations will find springs. What the statistic cannot do is predict
the consequence of an edit. It is a selector with no dynamics, and every row in
Table~\ref{tab:controls} uses FORCE to roll the edit forward;
the rows differ only in which pairs are selected.

We train on five-step rollouts; a sweep of training horizons shows longer
rollouts lower removal on both exact states and video, so the choice is safe
(Appendix~\ref{app:trainhorizon}).

\subsection{In-depth analysis}

\begin{wraptable}{r}{0.45\linewidth}
\centering
\vspace{-\baselineskip}
\caption{Ceiling analysis. Each row substitutes one more simulator quantity and
is a diagnostic; only the first row is a model result.}
\label{tab:ladder}
\vspace{2pt}
\small
\begin{tabular}{@{}lcc@{}}
\hline
substituted & $D_{\text{fixed}}$ & step \\
\hline
nothing (video)        & 0.1954 & --- \\
$+$ oracle selection   & 0.2481 & $+0.053$ \\
$+$ true velocity      & 0.3753 & $+0.127$ \\
$+$ true position      & 0.3476 & $-0.028$ \\
\hline
\end{tabular}
\end{wraptable}

Table~\ref{tab:ladder} substitutes simulator quantities one at a time. The start
velocity is the dominant recoverable term: supplying the true velocity after
oracle selection adds $0.127$, more than twice what perfect selection is worth.
Supplying the true position on top changes little ($-0.028$), so position is
not where the gap lies. Nor is the velocity term an estimator problem: the start
velocity the rollouts use, differenced from smoothed positions, is off by 65\% of
the true magnitude, and no estimator we tried recovers it from the perception
stream meaningfully better (Appendix~\ref{app:velocity}).

\begin{figure}[!h]
\centering
\begin{minipage}[t]{0.48\linewidth}
\centering
\includegraphics[width=\linewidth]{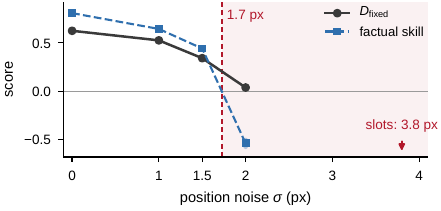}%
\vspace{-8pt}
\caption{Precision sweep on true positions plus $\sigma$ px of noise, two seeds:
$D_{\text{fixed}}$ and factual skill, with the zero-crossing (1.725 px) and the
measured centroid error (3.80 px) marked. A diagnostic.}
\label{fig:cliff}
\end{minipage}\hfill
\begin{minipage}[t]{0.48\linewidth}
\centering
\includegraphics[width=\linewidth]{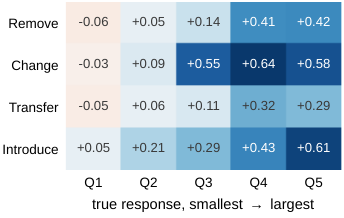}%
\vspace{-8pt}
\caption{$D_{\text{fixed}}$ by quintile of the true response magnitude, one
row per edit. Bins are defined by the simulator's response and are
model-independent.}
\label{fig:quintiles}
\end{minipage}
\end{figure}

\label{sec:cliff}
Training and scoring on true positions with Gaussian noise of $\sigma$ px
isolates the precision requirement from everything else
(Figure~\ref{fig:cliff}), and the result is a cliff, not a slope. The drop per
pixel is $-0.10$ from 0 to 1, $-0.37$ from 1 to 1.5 and $-0.61$ from 1.5 to 2.
Factual skill crosses zero at $\sigma = 1.725$ px (CI $[1.715, 1.736]$): beyond
that noise level the learned dynamics are \emph{worse than assuming free flight},
a regime change rather than degradation. Discovery collapses at the same point,
from $0.955$ to $0.564$, so this is not the edit failing against intact
selection; both fail together. The crossing is unchanged over a longer rollout of 21
steps ($1.728$ px), so the precision requirement is a property of how objects are located, not of
the horizon at which edits are scored. The measured centroid error of the
perception stream is 3.80 px, already past that threshold. The precision at
which relational editing becomes possible is therefore close to the precision
at which object-centric perception currently operates, and the video number
reflects that proximity.

Splitting scored entries into quintiles of the \emph{true} response magnitude,
so the bins are model-independent, shows the score rising steeply with the true
response on every edit (Figure~\ref{fig:quintiles}): removal runs from $-0.0555$ in the smallest
quintile to $0.4223$ in the largest, and on three of the four edits the smallest
quintile is negative, so the predicted response points the wrong way. The limit
is precision, not propagation: the model is right about large edits and wrong
about small ones.

The same mechanism explains why restricting the metric to the two edited objects
raises every edit, and why transfer is the weakest edit
(Appendix~\ref{app:editedpair}).

We score at $H=10$, selected on the development fixture from a sweep of
scoring horizons; the held-out clips test the choice, and development and
held-out removal agree to within $0.007$ (Appendix~\ref{sec:horizon}).

\section{Limitations}
\label{sec:limitations}

We study relations that persist, such as a spring that acts on its pair
throughout a clip, and the held relation latent is designed for that case.
Relations that act only briefly, such as collisions, call for a different form:
properties of each object combined with their geometry at the moment of
contact, rather than a relation read from a window and held; we leave this to
future work.

The interaction head assumes a force along the line between two objects, with a
fixed expression for its strength. On exact object states, replacing that
expression with a learned function of the relation latent and the distance
improves every metric, although it trains less reliably, so the fixed form does
not limit the results here. Forces off the line between centres are outside what
the head can express. The integrator and the form of the force also match the
simulator's, a structural prior shared with the environment even though no
parameter is fitted to ground truth.

Perception sets the ceiling. Editing needs objects located more precisely than
current object-centric perception achieves, and the velocity at the start of a
rollout is where the shortfall matters most. Better estimation of object motion
from video is the clearest route to better counterfactuals, and nothing in FORCE's
relation encoder or interaction head needs to change to benefit from it.
Scoring counterfactuals in pixel space rather than in state is left open
(Appendix~\ref{app:decoder}).

\section{Conclusion}
\label{sec:conclusion}

We presented FORCE, a world model in which each pairwise interaction is an
explicit latent, inferred once and held fixed, so that removing, changing,
transferring or introducing a relation is a direct write to the model's own
quantities. Scored by the change each edit induces against the simulator's,
all four edits clear their controls, and the same architecture trained on exact
positions edits far more accurately, which places the remaining gap in how precisely
objects are located from video, not in the relational dynamics. Better
object-centric perception benefits FORCE directly, and relations that act only
briefly, such as contact, are the natural next step.

\section*{Reproducibility statement}

The environment, its parameters and the split by simulation are described in
Section~\ref{sec:experiments} and Appendix~\ref{app:env}; the perception
objective, with every term and weight, in Appendix~\ref{app:stage0}; the relation
training settings in Appendix~\ref{app:relations}; and both training and the
application and scoring of an edit as algorithms in Appendix~\ref{app:algorithms}.
Every reported number is five seeds unless stated, and the evaluation protocol,
controls, selection rule and held-out clips are specified in
Section~\ref{sec:experiments}. Code will be made publicly available.

\section*{AI use statement}

Generative AI tools were used for limited coding and writing assistance. For coding, these tools were primarily used for minor code edits and autocomplete support during the implementation and analysis of experiments. For writing, they were used to assist with drafting portions of the manuscript and improving grammar, phrasing, and overall readability. All generated or suggested content was reviewed and verified by the authors, who take full responsibility for the final manuscript.

\bibliography{iclr2027_conference}
\bibliographystyle{iclr2027_conference}

\appendix
\raggedbottom

\section{Environment}
\label{app:env}

Four objects of radius $0.08$ move in a unit box under two springs, drawn
independently per simulation and fixed for its duration. The spring constant is
$10.0$; rest lengths are drawn from $\mathcal{N}(0.41, 0.031)$ and clipped to
$[0.317, 0.503]$ on their own random stream; speed is capped at $0.024$ and there
is no damping. The development dataset is 400 simulations of 3000 frames, split
by simulation into 280 for training, 40 for validation and 80 for testing; the
80 held-out simulations are generated separately with the same settings. The
cached perception stream used in training covers 150 of the training
simulations and none of the validation or test simulations.

Objects are drawn as distinct shapes rather than plain discs because frozen
DINOv2 features distinguish objects mainly by form: with discs differing only in
colour, slot attention did not bind to individual objects. Shape and colour are
drawn independently for each simulation, so appearance identifies an object
within a clip but carries no information about which pairs are coupled. Each
object is drawn as a solid silhouette chosen from ten shapes (circle,
square, triangle, diamond, cross, ring, pentagon, bowtie, hexagon, star) in one
of ten colours, with shape and colour drawn independently per simulation,
giving 100 possible appearances. Shape affects rendering only: the dynamics
treat every object as a disc of radius $0.08$, and forces act between centres.

\section{Perception training}
\label{app:stage0}

The slot model is trained once, in a first stage, and frozen before any relation
is learned. It reads frozen DINOv2-S/14 patch features \citep{oquab2023} from
$224\times224$ frames, on a $16\times16$ grid of 384-dimensional features, with
$K=5$ slots and 10 slot-attention iterations \citep{locatello2020}, over sequences
of 16 frames sampled at a stride of 3. In the dynamics model that predicts slots
forward, attention between slots is restricted so that each slot attends only to
its own history. Training runs for 4000 steps with Adam, the learning rate
warming up to $2\times10^{-4}$ over the first 500 steps and then decaying on a
cosine schedule to $10^{-4}$. The slots are then frozen at a
learning rate of exactly zero; the configuration refuses any nonzero value.
No term reconstructs pixels: every target is a frozen DINOv2 feature map, following
\citet{seitzer2023}.

\begin{table}[H]
\centering
\caption{Terms of the perception objective.}
\label{tab:stage0}
\vspace{4pt}
\small
\begin{tabular}{@{}p{0.22\linewidth}p{0.48\linewidth}p{0.24\linewidth}@{}}
\hline
term & what it compares & weight \\
\hline
next-step prediction & predicted against actual DINOv2 features of the next
frame, weighted towards moving regions & 1.0, motion weight 2.0 \\
rollout & slots predicted over 7 steps from frame 7 against slots inferred from
the frames & 1.0 \\
feature reconstruction & a broadcast decoder's reconstruction of the DINOv2
feature map, with per-slot masks normalised across slots & 0 for 250 steps, then
linear to 0.5 over 750 \\
latent regulariser & KL divergence on the slot latents & ramped from
$10^{-6}$ to $2\times10^{-4}$ \\
relation terms & all terms involving relations & 0 in this stage \\
\hline
\end{tabular}
\end{table}

The checkpoint used throughout was trained with seed 11. After training, each
slot's attention centroid and pooled optical flow are cached once for the whole
dataset. Without labels, the background slot is identified by its attention mass
and the remaining slots are taken in index order.

\section{Relation training}
\label{app:relations}

The relation encoder and interaction head are trained jointly for 256k steps at a
batch of 64, using Adam at a learning rate of $10^{-3}$ with default moments and
no weight decay or schedule, and with the gradient norm clipped at 1.0 at every
step. The relation latent has width 8, and the training rollout is 5 steps and
free-running from the first step, as described in Section~\ref{sec:method}.

Two settings differ between training and evaluation and are stated here rather
than left implicit. The rollout horizon is short during training and longer at
evaluation, and the evaluation stream uses positions smoothed by a fixed-interval
smoother \citep{rauch1965} with one-frame differenced velocities, where training
uses raw cached centroids with flow velocities; a second optical-flow estimator
\citep{kroeger2016dis} sets the smoother's velocity noise without labels.

\section{Algorithms}
\label{app:algorithms}

Algorithm~\ref{alg:train} gives the training of the relation encoder and
interaction head, and Algorithm~\ref{alg:edit} gives how an edit is applied and
scored. In both, the relation latent is inferred once from the observation window
and held for the whole rollout.

\begin{algorithm}[H]
\caption{Training the relation encoder and interaction head.}
\label{alg:train}
\begin{algorithmic}[1]
\Require cached positions $x$ and flow velocities $v$ for the training clips;
window $W=16$; training horizon $H=5$
\For{256k steps}
  \State sample a batch of clips, each with window frames $t_0-W+1,\dots,t_0$ and
  targets $t_0+1,\dots,t_0+H$
  \State $\phi_{ij} \gets \big[\lVert d_{ij}\rVert,\ \lVert d_{ij}\rVert^{2},\
  (v_j - v_i)\cdot\hat u_{ij}\big]$ over the window, for every ordered pair
  \State $\mathbf m \gets \mathrm{RelationEncoder}(\phi)$;\quad
  $\mathbf m \gets \tfrac12(\mathbf m + \mathbf m^{\top})$
  \Comment{inferred once, then held}
  \State $(\alpha, \beta) \gets U\mathbf m + c$
  \Comment{interaction head}
  \State $\hat x \gets x_{t_0}$;\ $\hat v \gets v_{t_0}$;\ $\mathcal L \gets 0$
  \For{$h = 1, \dots, H$}
    \State $f_{ij} \gets (\alpha_{ij}\lVert d_{ij}\rVert - \beta_{ij})\,\hat u_{ij}$
    for every pair, with $d_{ij}$ from the current $\hat x$
    \State $a_i \gets \sum_{j \neq i} f_{ij}$
    \State $(\hat x, \hat v) \gets \mathrm{Integrator}(\hat x, \hat v, a)$
    \Comment{the simulator's update rule}
    \State $\mathcal L \gets \mathcal L + \lVert \hat x - x_{t_0+h} \rVert^{2}$
    \Comment{observations are targets only}
  \EndFor
  \State Adam step on $\mathcal L / H$, learning rate $10^{-3}$, gradient norm
  $\le 1$
\EndFor
\end{algorithmic}
\end{algorithm}

\begin{algorithm}[H]
\caption{Applying and scoring an edit.}
\label{alg:edit}
\begin{algorithmic}[1]
\Require a test clip with anchor $t_0$; an edit request naming an operation, a
target pair $(i,j)$, and a donor pair $(k,l)$ or a shift $\delta$; scoring
horizon $H=10$; the simulator's stored state $s^{*}_{t_0}$
\Statex \emph{Model}
\State $(\alpha, \beta) \gets$ coefficients for every pair, from the window
\Comment{as in Algorithm~\ref{alg:train}}
\State $\text{keep} \gets \mathrm{Otsu}\big(\sqrt{\alpha^{2}+\beta^{2}}\big)$;\quad
$(\alpha, \beta) \gets (\alpha, \beta) \odot \text{keep}$
\Comment{reads no adjacency}
\State $(\alpha', \beta') \gets (\alpha, \beta)$, then apply the edit:
\Statex \hspace{\algorithmicindent} Remove: $\alpha'_{ij} = \beta'_{ij} = 0$
\qquad Change: $\beta'_{ij} = \beta_{ij} + \delta\,\alpha_{ij}$
\Statex \hspace{\algorithmicindent} Transfer: $\beta'_{ij} = \alpha_{ij}\,\beta_{kl}/\alpha_{kl}$
\qquad Introduce: $(\alpha'_{ij}, \beta'_{ij}) = (\alpha_{kl}, \beta_{kl})$
\State $\Delta\hat y \gets \mathrm{Roll}(\alpha', \beta', H) - \mathrm{Roll}(\alpha, \beta, H)$
\Comment{same start state}
\Statex \emph{Simulator}
\State $\Delta y^{*} \gets \mathrm{Sim}(s^{*}_{t_0}, H, \text{edited}) - \mathrm{Sim}(s^{*}_{t_0}, H)$
\Comment{same edit, true springs}
\Statex \emph{Score, per entity $e$ at step $H$}
\State match entities to the simulator's objects by position
\Comment{permutes indices only}
\State $\mathcal E \gets \{\, e : \lVert \Delta y^{*}_e \rVert > 0 \,\}$
\Comment{entities the edit truly moves}
\State $s_e \gets \cos(\Delta\hat y_e, \Delta y^{*}_e)$ if
$\lVert \Delta\hat y_e \rVert > 0$, otherwise $s_e \gets 0$
\State \Return $D_{\text{fixed}} = \frac{1}{|\mathcal E|}\sum_{e \in \mathcal E} s_e$,
averaged over clips
\end{algorithmic}
\end{algorithm}

\section{Estimating the start velocity}
\label{app:velocity}

The start velocity is the largest term in the ceiling analysis, so we tested
whether a better estimator could recover it from the perception stream. Out of
sample, the start-velocity error, as a fraction of the true magnitude, is
$64.8\%$ for the velocity the scored rollouts use, differenced from smoothed
positions; $66.7\%$ from optical flow pooled under each slot's attention map;
$81.1\%$ from the whole 16-frame position track; and $99.0\%$ from slot vectors,
against $100\%$ for predicting zero. No combination improves on the first two,
and three relation models trained with the alternatives in place of flow all
scored lower. The ceiling analysis substitutes the true velocity for the first
of these. The information needed to seed a rollout accurately is not present in
the perception stream, whatever reads it.

\section{Edited-pair metric and transfer}
\label{app:editedpair}

The mechanism behind the quintile trend (Section~\ref{sec:cliff}) explains a second observation. Restricting the metric to the
two edited objects rather than all four raises every edit (removal $0.1954 \to
0.3211$, change $0.3650 \to 0.5743$). The entries the full metric adds are third
objects reached only by propagation, and those carry small true responses,
exactly where Figure~\ref{fig:quintiles} shows the model is weakest. These are one
phenomenon seen twice, not two results. It also accounts for transfer, the
weakest edit: a transfer gives a pair another pair's rest length, so its true
response is set by the difference between the two, which is usually small, and
the whole transfer distribution lies in the range where every edit is hard.
About one transfer in eight also names a donor pair that was not selected, in
which case the written value is zero.

\section{Scoring horizon}
\label{sec:horizon}

\begin{figure}[H]
\centering
\includegraphics[width=0.55\linewidth]{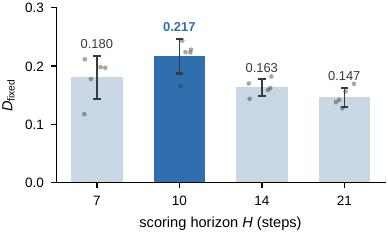}%
\vspace{-8pt}
\caption{$D_{\text{fixed}}$ against the scoring horizon on a matched clip set,
five seeds (dots). Coverage is 0.83 at every horizon, so the differences are in
quality. The horizon used, $H=10$, is highlighted.}
\label{fig:horizon}
\end{figure}

Sweeping the horizon on a matched clip set shows $D_{\text{fixed}}$ rising to a
peak near $H=10$ and declining at longer horizons (Figure~\ref{fig:horizon}), a
pattern that replicates on an independent clip set. Coverage saturates within the
first few steps, so the effect is in quality: the predicted and true responses
agree best at intermediate horizons, consistent with a small error in the model's
estimate of the oscillation accumulating over the rollout. We therefore score at
$H=10$. Because the choice was made on the development fixture, the held-out
clips test it, and development and held-out removal agree to within $0.007$.

\section{Training horizon}
\label{app:trainhorizon}

\begin{figure}[H]
\centering
\includegraphics[width=\linewidth]{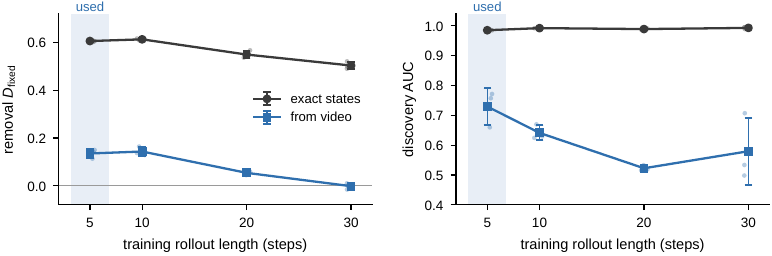}%
\vspace{-8pt}
\caption{Effect of the training rollout length at 64k steps, three seeds per
horizon (dots; bars show $\pm$ SD). Left: removal $D_{\text{fixed}}$ at the
scoring horizon against training horizon, on exact states and from video.
Right: discovery AUC. The horizon used in the paper, five steps, is shaded.}
\label{fig:trainhorizon}
\end{figure}

We train on five-step rollouts, and a sweep of training horizons shows the
choice is safe. Training on longer rollouts of 20 or 30 steps lowers removal on
both exact states and video, while rollouts of 5 and 10 steps end within $0.007$
of each other on exact states (Figure~\ref{fig:trainhorizon}). From video, longer rollouts also cost
discovery, which falls from AUC $0.73$ at 5 steps to $0.52$ at 20 while staying
near $0.99$ on exact states. The sweep uses three seeds per horizon and runs to
64k steps, against 256k for the reported models.

\section{Rendering}
\label{app:decoder}

The decoder renders predicted states back to frames, so that the pipeline runs
from pixels to a counterfactual and back to pixels. It is used for figures only;
no quantity reported in this paper passes through it.

The decoder is a conditional diffusion model \citep{ho2020ddpm} that takes the
entity slots at the start of the rollout, held fixed, together with the predicted
positions and a reference frame. It is trained with the velocity
parameterisation \citep{salimans2022} on a cosine noise schedule
\citep{nichol2021} and sampled deterministically \citep{song2021ddim}. The same
initial noise is used for the unedited and edited renders of a clip, so the two
differ only through the predicted positions.

Our evaluation scores predicted states rather than rendered frames. Scoring
counterfactuals in pixel space, where a correct edit must also look right, is
the natural complement, and would need a decoder that preserves object
identity over long rollouts and a metric that separates the effect of an edit
from ordinary rendering error.

\end{document}

%% file: math_commands.tex
\usepackage{amsmath,amsfonts,bm}

\def\eqref#1{equation~\ref{#1}}
\def\1{\bm{1}}

\DeclareMathAlphabet{\mathsfit}{\encodingdefault}{\sfdefault}{m}{sl}
\SetMathAlphabet{\mathsfit}{bold}{\encodingdefault}{\sfdefault}{bx}{n}